\documentclass{IEEEtran}
\IEEEoverridecommandlockouts
\usepackage{cite}
\usepackage{amsmath,amssymb,amsfonts}
\usepackage{algorithmic}
\usepackage{graphicx}
\usepackage{textcomp}
\usepackage{xcolor}
\usepackage{nkj}

\usepackage{verbatim}
 \graphicspath{{../}}

\usepackage{algorithm}
\usepackage{algorithmic}

\usepackage{subfigure}
\newif\ifReview
\Reviewtrue

\newif\ifOmitAppendix
\OmitAppendixtrue

\newif\ifSuppressMemo
\ifSuppressMemo
\newcommand{\memo}[1]{}
\else
\usepackage{color}
\newcommand{\memo}[1]{{\bf \textcolor{red}{[#1]}}}
\fi

\newcommand{\method}{L\'evy-Attack}

\def\BibTeX{{\rm B\kern-.05em{\sc i\kern-.025em b}\kern-.08em
    T\kern-.1667em\lower.7ex\hbox{E}\kern-.125emX}}
\begin{document}

\title{Black-Box Decision based Adversarial Attack with Symmetric $\alpha$-stable Distribution\\
\thanks{This work was supported by
the German Ministry for Education and Research as Berlin Big Data Center BBDC (funding mark 01IS18025A) and Berlin Center for Machine Learning BZML (funding mark 01IS18037I). The work of K.-R. M\"{u}̈ller was supported by the Institute for Information and Communications Technology Promotion Grant funded by the Korea government (MSIT) (No. 2017-00451, No. 2017-0-01779).
}
}

\author{
 Vignesh Srinivasan$^{1}$, Ercan E. Kuruoglu$^{2}$, Klaus-Robert M{\"u}ller$^{3,4,5,6}$\\
{  {Wojciech Samek$^{1,4}$
and Shinichi Nakajima$^{3,4,7}$} }\\
$^1$Fraunhofer HHI,
$^2$Instititute of Science and Technologies of Information, CNR, Pisa, Italy, \\
$^3$TU Berlin,
$^4$Berlin Big Data Center,
$^5$Korea University,
$^6$MPI for Informatics,
$^7$RIKEN AIP\\ 
\texttt{\{vignesh.srinivasan,wojciech.samek\}@hhi.fraunhofer.de}\\
\texttt{\{ercan.kuruoglu\}@isti.cnr.it}\\
\texttt{\{klaus-robert.mueller,nakajima\}@tu-berlin.de }\\
}


\maketitle

\begin{abstract}
Developing techniques for adversarial attack and defense is an important research field
for establishing reliable machine learning and its applications.
Many existing methods employ Gaussian random variables for exploring
the data space to find the most adversarial (for attacking)
or least adversarial (for defense) point.
However, the Gaussian distribution is not necessarily the optimal choice when the exploration is required to follow the complicated structure that most real-world data distributions exhibit.
In this paper, we investigate how statistics of random variables affect 
such
random walk exploration.
Specifically, we generalize the \emph{Boundary Attack}, a state-of-the-art 
black-box decision based attacking strategy,
and propose the \emph{\method{}},
where the random walk is driven by symmetric $\alpha$-stable random variables.
Our experiments on MNIST and CIFAR10 datasets show that the \method{}
explores the image data space more efficiently,
and significantly improves the performance.
Our results also give an insight into the recently found fact in the whitebox attacking scenario that the choice of the norm for measuring the amplitude of the adversarial patterns is essential.
\end{abstract}


\section{Introduction}
\label{sec:Introduction}
The success of deep neural networks (DNNs) \cite{Krizhevsky,lenet,googlenet,vgg,resnet}
has led to them being used in many real world applications. 
However, these models are also known to be susceptible to adversarial attacks, i.e.,
minimal patterns crafted by attackers who try to fool learning machines
\cite{Goodfellow, Papernotb, szegedy, Nguyena, Eykholt,athalye3d}. Such adversarial patterns
do not affect human perception much, while they can manipulate learning machines, e.g., to give wrong classification outputs.
DNN's complex interactions between different layers enable high accuracy under the controlled setting, while they make the outputs unpredictable in \emph{untrained spots} where training samples exist sparsely.
If attackers can find such a spot close to a normal data sample,
they can manipulate DNNs by adding a very small (optimally invisible in computer vision applications) perturbation to the original sample,
leading to fatal errors,
e.g., manipulating an autonomous driving system can cause serious accidents.

Two attacking scenarios are considered in general---whitebox and blackbox. 
The whitebox scenario assumes that the attacker has access to the complete target system, including the architecture and the weights of the DNN, as well as the defense strategy if the system is equipped with any. 
Typical whitebox attacks optimize the classification output with respect to the input by backpropagating through the defended classifier
\cite{Carlinib, chen2017ead, sharma2017ead, moosavi2016deepfool}.
On the other hand, the blackbox scenario assumes that the attacker has only access to the output.  Under this scenario, the attacker has to rely on blackbox optimization, where the objective can be computed for arbitrary inputs, but the gradient information is not directly accessible.
Although the whitebox attack
is more powerful,
it is much less likely 
that attackers can get full knowledge of the target system in reality. 
Accordingly, the blackbox scenario 
is considered to be a more realistic 
threat.

Existing blackbox attacks can be classified into two types---the transfer attack and the decision based attack.
In the transfer attack, the attacker trains a student network which mimics the output of the target classifier.
The trained student network is then used to get the gradient information for optimizing the adversarial input.
In the decision based attack,
the attacker simply
performs random walk exploration.
In the \emph{boundary attack}  \cite{brendel2017decision}, a state-of-the-art method in this category,
the attacker first generates an initial adversarial sample from a given original sample
by drawing a uniformly distributed random pattern multiple times
until it happens to lead to misclassification.
Initial patterns generated in this way typically have too large amplitudes to be hidden from human perception.
The attacker therefore polishes the initial adversarial pattern by Gaussian random walk in order to minimize the amplitude, keeping the classification output constant.%
\footnote{In the case of the untargeted attack, the classification output is kept \emph{wrong}, i.e., random walk can go through the areas of any label except the true one.}

Here our question arises.  Is the Gaussian appropriate to drive the adversarial pattern to minimize the amplitude?
It could be a reasonable choice if we only consider that the attacker minimizes the $L_2$ norm of the adversarial pattern.
However, it is also required to keep the classification output constant through the whole random walk sequence.
Provided that the decision boundary of the classifier has complicated structure, reflecting the real-world data distribution,
we expect that more efficient random walk can exist.

In this paper, we pursue this possibility, and 
investigate how statistics of random variables affect the performance of attacking strategies.
To this end, 
we generalize the boundary attack,
and propose the \method{} where 
the random walk exploration is 
driven by symmetric $\alpha$-stable random variables.
We expect that 
 the impulsive characteristic of the $\alpha$-stable distribution induces sparsity in random walk steps, 
 which would drive adversarial patterns along the complicated decision boundary structure efficiently.
 Naturally, our expectation is reasonable only if the decision boundary has some structure aligned to the coordinate system defined in the data space,
 so that moving along the canonical direction keep more likely the classification output than moving isotropic directions. 

In our experiments
on MNIST and CIFAR10 datasets,
\method{} with $\alpha \sim 1.0$ or less shows significantly better performance than
the original boundary attack with Gaussian random walk.
This implies that our hypothesis on the decision boundary holds at least in those two popular image benchmark datasets.
Our results also give an insight into the recently found fact in the whitebox attacking scenario that the choice of the norm for measuring the amplitude of the adversarial patterns is essential.




\section{Proposed Method}

In this section, we first introduce the $\alpha$-stable distribution,
and propose the \method{} as a generalization of the boundary attack.

\subsection{Symmetric $\alpha$-stable Distribution}
\label{sec:alpha_stable}

The symmetric $\alpha$-stable distribution is a generalization of the Gaussian distribution which can model characteristics too impulsive for the Gaussian model. This family of distributions
is most conveniently defined by their characteristic functions \cite{samorodnitsky94} due to the lack of an analytical expression for the probability density function.
The characteristic function is given as
\begin{equation}
\phi (s) = \exp [i\mu s - \left| {\gamma s} \right|^\alpha ],
\end{equation}
where $\alpha \in (0,2],   \mu \in (-\infty,\infty)$, and $\gamma \in (0,\infty)$ are parameters.
 We denote the $D$-dimensional symmetric $\alpha$-stable distribution by $\mathcal{SA}_D(\alpha, \mu, \gamma)$.
$\alpha $ is the characteristic exponent expressing 
 the degree of \emph{impulsiveness} of  the distribution---%
 the smaller $\alpha $ is, the more impulsive the distribution is.
The  symmetric $\alpha$-stable distribution reduces to the Gaussian distribution for $\alpha  = 2$,
and to the Cauchy distribution for $\alpha = 1$, respectively.
$\mu$  is the location parameter, which  corresponds to the mean in the Gaussian case,
while
$\gamma$ is the scale parameter measuring of the spread of the samples around the mean, which corresponds to the variance in the Gaussian case. 
For more details on $\alpha$-stable distributions, readers are referred to \cite{samorodnitsky94}.





\begin{algorithm}[t]
  \begin{algorithmic}[1]
  \item
\begin{flushleft}
\textbf{Input:}
Classifier $\bff(\cdot)$, original image $\bfx$ and label $\bfy$
Max. number $T$ of iterations, termination threshold $\psi$ \\
\textbf{Output:} Adversarial sample $\bfx^-$
\end{flushleft}
 \REPEAT  
   \STATE      $\bfx^{-}_{0} \leftarrow \bfx + {\bf\Delta}$ for ${\bf\Delta} \sim  \mathcal{U}_D (0,255)$ 
\UNTIL{$\bfy \neq  \bff(\bfx^{-}_{0}) $}
\FOR {$t = 0$ to $T-1$}
	\STATE $(\bfx^{-}_{t+1}, \epsilon) \leftarrow \textit{$\alpha$-stable random update}(\bfx^{-}_{t})$ 
	\IF{$\bfy =  \bff(\bfx^{-}_{0}) $} 
    \STATE $\bfx^{-}_{t+1} \leftarrow \bfx^{-}_{t}$
    \ENDIF 
	\IF { $\epsilon < \psi$}
	   \item \textbf{break}
	\ENDIF
\ENDFOR
\end{algorithmic}
\caption{(Untargeted) \method{} }
\label{alg:LevyAttack}
  \vspace{0mm}
\end{algorithm}

\subsection{\method{}}
\label{sec:boundary_attack}

Now, we propose our \method{} as a generalization of the boundary attack \cite{brendel2017decision},
 a simple yet effective attack under the blackbox scenario, where the attacker has access only to the classification output.
We denote the classifier output by $\bfy = \bff(\bfx)$, where 
$\bfx$ is a data point, and $\bff$ is the target (blackbox) classifier.
 The \method{} performs the procedure as described in Algorithm~\ref{alg:LevyAttack}, which reduces to the original boundary attack if we set $\alpha=2$. 
 
  While the attack is very simple, it can be a very effective blackbox attack \cite{brendel2017decision}. 
Naturally, 
the success of the \method{} relies on the effectiveness of the proposal distribution. 
In the proposal distribution we accommodate sampling from a symmetric $\alpha$-stable distribution  $\bfeta_t \sim \mathcal{SA}_D(\alpha,0,1)$. 
$\| \bfeta_t \|_2 = \delta \cdot d(\bfx^{-}_{t}, \bfx)$, where $d(\bfx^{-}_{t}, \bfx) = \| \bfx^{-}_{t+1} - \bfx \|^2_2$ and $\delta$ is the relative size of the perturbation. 
An orthogonal step is taken, where $\bfeta_t$ is projected onto a sphere around the original image such that $d(\bfx^{-}_{t}, \bfx) = d(\bfx^{-}_{t+1}, \bfx)$.
Finally, a step is taken towards the original image so that the adversarial perturbation is reduced by a small amount $\epsilon$,      $d(\bfx^{-}_{t}, \bfx) - d(\bfx^{-}_{t+1}, \bfx) = \epsilon \cdot d(\bfx^{-}_{t}, \bfx)$.

$\delta$ and $\epsilon$ are the two hyper-parameters which are dynamically tuned to adjust to the local geometry of the decision boundary. 
The orthogonal step in the proposal distribution encourages around $50\%$ orthogonal perturbations to be adversarial. 
The length of the step, $\epsilon$ is adjusted according to the success rate of the step. If the success rate is too high, then $\epsilon$ is increased and vice versa. 
As the random walk moves closer towards the original image, the success rate of the attack becomes lesser. The attack gives up further exploration as $\epsilon$ converges to $0$.

\begin{table}[t]
\caption{The mean $  \mathcal{S}_{m}$ and the median $  \mathcal{S}_{d}$ of the $L_\infty$, $L_1$, and $L_2$ norms of the adversarial patterns generated by \method{} for the MNIST dataset.}
\begin{center}
\begin{tabular}{|c|c|c|c|c|c|c|}
\hline

\textbf{Attack} & \multicolumn{2}{c}{\textbf{$L_\infty$}} & \multicolumn{2}{c}{\textbf{$L_1$}} & \multicolumn{2}{c}{\textbf{$L_2$}} \\
\cline{2-7} 

& $\mathcal{S}_{m}$ & $\mathcal{S}_{d}$ & $\mathcal{S}_{m}$ & $\mathcal{S}_{d}$ & $\mathcal{S}_{m}$ & $\mathcal{S}_{d}$ \\

\hline
Gaussian       &    \textbf{0.56} & \textbf{0.56} & 11.36 & 10.73 & 1.38 & 1.39 \\
\hline
$\alpha = 1.5$ &    0.57 & 0.58 & 9.62  & 9.16  & 1.31 & 1.31\\
\hline
$\alpha = 1.0$ &    0.57 & 0.58 & 8.89  & \textbf{8.54}  & \textbf{1.29} & \textbf{1.30} \\
\hline
$\alpha = 0.5$ &    0.58 & 0.58 & \textbf{8.84}  & 8.71  & 1.30 & 1.32   \\
\hline
\end{tabular}
\label{table:mnist}
\end{center}
\end{table}

\begin{figure}[t]
    \centering
    \centering
    \includegraphics[width=\linewidth]{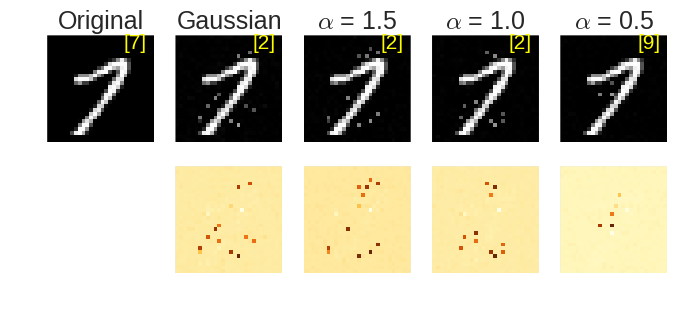}
    \centering
    \includegraphics[width=\linewidth]{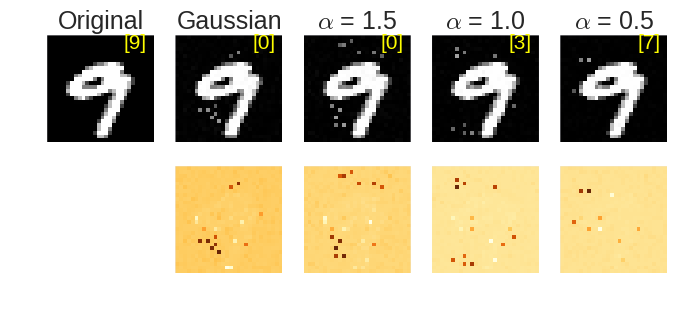}
    
\caption{Adversarial samples generated by \method{} on MNIST dataset for "7" and "9".  
"Gaussian" corresponds to $\alpha = 2$, with which  \method{} reduces to the original boundary attack  \cite{brendel2017decision}.
The classification output is shown at the top right corner of each image. In each block (for "7" as well as "9"), the top row displays adversarial samples generated with different $\alpha$, while the bottom row displays the corresponding adversarial patterns (the differences from the original image). }
\label{fig:mnist}
\end{figure}

\section{Experiment}
We report on experiments performed using our \method{} on the following datasets:
\begin{itemize}
    \item MNIST: The MNIST dataset consists of $60,000$ images in total, with $50,000$ images for training and $10,000$ images for testing. It has $10$ different classes each corresponding to the $10$ numerical digits. The image size is $28 \times 28$. 
    
    \item CIFAR10: This dataset also contains $50,000$ training images and $10,000$ test images. The images are of resolution $32 \times 32 \times 3$ with $10$ different classes in total.  
     
\end{itemize}


In the MNIST experiment, we target the state-of-the-art robust classifier proposed by Madry et al. \cite{madry2017towards},%
\footnote{https://github.com/MadryLab/mnist\_challenge}
where the classifier is trained,  in addition to the original training set, on the adversarial samples generated by the PGD attack of bounded $L_\infty$ distortion by $0.3$. 
The classification accuracy on the original test samples is $98.68\%$.
In the CIFAR10 experiment, we trained a state-of-the-art Resnet model \cite{resnet}. 
The classification accuracy on the original test samples is $92.93\%$.
    
 To generate samples by \method{},
 we
    modify the code
    provided by \cite{brendel2017decision} for the boundary attack,
    so that the random walk is performed by
    the symmetric $\alpha$-stable distribution, instead of the Gaussian distribution. 
We evaluated adversarial samples for
$\alpha =2.0, 1.5, 1.0$, and $0.5$.
The other parameters specifying the $\alpha$-stable distribution is set to  $\delta = 0.0$ and $\gamma = 1.0$.
We limit the number of random walk steps to $5,000$.
Having such an upper-bound is reasonable
because it is not realistic to assume that  
the attacker may access to the classifier output unlimited times.

\begin{table}[t]
\caption{The mean $  \mathcal{S}_{m}$ and the median $  \mathcal{S}_{d}$ of the $L_\infty$, $L_1$, and $L_2$ norms of the adversarial patterns generated by \method{} for the CIFAR10 dataset.}
\begin{center}
\begin{tabular}{|c|c|c|c|c|c|c|}
\hline

\textbf{Attack} & \multicolumn{2}{c}{\textbf{$L_\infty$}} & \multicolumn{2}{c}{\textbf{$L_1$}} & \multicolumn{2}{c}{\textbf{$L_2$}} \\
\cline{2-7} 

& $\mathcal{S}_{m}$ & $\mathcal{S}_{d}$ & $\mathcal{S}_{m}$ & $\mathcal{S}_{d}$ & $\mathcal{S}_{m}$ & $\mathcal{S}_{d}$ \\

\hline
Gaussian       &    \textbf{2.92} & 2.47 & 895.22 & 755.06 & 23.72 & 20.45 \\
\hline
$\alpha = 1.5$ &    2.99 & 2.44 & 859.49 & 708.54 & 23.15  & 19.49 \\
\hline
$\alpha = 1.0$ &    2.97 & 2.427 & 847.20 & 700.42  & 23.06  & 19.39 \\
\hline
$\alpha = 0.5$ &    2.94 & \textbf{2.421} & \textbf{826.29} & \textbf{685.76}  & \textbf{22.78}  & \textbf{19.28}   \\
\hline
\end{tabular}
\label{table:cifar}
\end{center}
\end{table}

\begin{figure}[t]
    \centering
    \includegraphics[width=\linewidth]{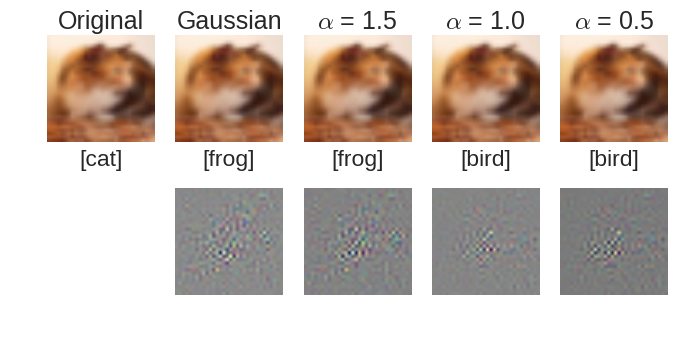}
    \includegraphics[width=\linewidth]{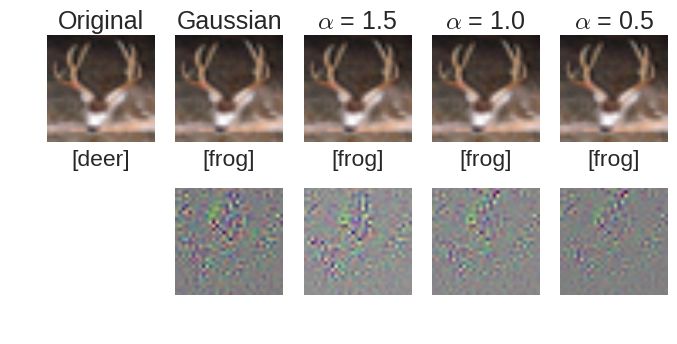}
    
\caption{Adversarial samples generated by \method{} on CIFAR10 dataset for "cat" and "deer".  
}
\label{fig:cifar}
\end{figure}

%
%

\begin{table}
\caption{The average number of iterations that \method{} performed to generate adversarial samples.}
\begin{center}
\begin{tabular}{|c|c|c|}
\hline

\textbf{Attack} & MNIST & CIFAR10 \\

\hline
Gaussian       &    2700.22  & 4996.49 \\
\hline
$\alpha = 1.5$ &    2629.04 & 4995.96 \\
\hline
$\alpha = 1.0$ &    2792.52 & 4987.04  \\
\hline
$\alpha = 0.5$ &    3407.54 & 4997.37  \\
\hline
\end{tabular}
\label{table:iterations}
\end{center}
\end{table}

For both datasets, we randomly sample $N=1,000$ images from the test set,
 and evaluate the quality of the adversarial patterns. 
As evaluation scores,
we use the mean and the median of $3$ different $L_p$-norms for $p = \infty, 1$, and $2$, over the 1,000 samples:
\begin{align}
    \mathcal{S}_{m} &= \frac{1}{N} \sum^{N}_{i=1} ( \| \bftau_i \|_p ) ,
&    \mathcal{S}_{d} &= median_{i=1}^{N} ( \| \bftau_i \|_p ) ,
\notag
\end{align}
where $\{\bftau_i\}$ are the adversarial patterns.
Smaller norms indicate that the adversarial pattern is less visible,
and therefore a better attack.

Table~\ref{table:mnist} shows the results 
on the MNIST dataset, where we see that the \method{} with $\alpha$ smaller than 2 (Gaussian) 
gives significantly smaller $L_1$ and $L_2$ norms with the $L_{\infty}$ norm almost unchanged.
Similar results are obtained on the CIFAR10 dataset  (Table~\ref{table:cifar}),
where $\alpha < 2$ gives better $L_1$ and $L_2$ norms with the $L_{\infty}$ norm almost unchanged,
although the performance difference is smaller than the MNIST results. 

Table~\ref{table:iterations} summarizes the average number of iterations the \method{} performs. 
We see the tendency that smaller $\alpha$ leads to more iterations,
which implies that $\alpha$-stable random walk continues exploring when Gaussian random walk has already been terminated due to a low success rate in further adversarial exploration.
Also, Tables~\ref{table:mnist} and \ref{table:cifar} show that the \method{} for $\alpha < 2$ reaches to the point closer to the original than the Gaussian ($\alpha = 2$)
random walk.
These results imply that the impulsive random walk is suitable to explore the data space without crossing decision boundaries,
and
indirectly support our hypothesis in Section~\ref{sec:Introduction}---decision boundaries have some structure aligned to the coordinate system.


Figs.~\ref{fig:mnist} and \ref{fig:cifar} show a few illustrative examples of adversarial samples and adversarial patterns generated by \method{}.
For each block for the examples ("7" and "9" in MINST, and "cat" and "deer" in CIFAR10), 
the top row shows the generated adversarial samples, while the bottom row shows the corresponding adversarial patterns (the differences from the original image).
In the "7" example of MNIST (Fig.~\ref{fig:mnist}), \method{} with $\alpha \geq 1.0$ consistently tries to modify the sample close to "2",
while \method{} for $\alpha = 0.5$ tries to modify the sample close to "7".
Apparently, the latter is more efficient, i.e., it requires fewer pixels to make "7" to "9" than to make "7" to "2".
The same applies to the "9" example---it seems more efficient to make "9" close to "7" than to make "9" to "0" or "3".
However, only \method{} with a very small $\alpha$ can find those efficient solution,
because it is little likely to get a sparse random walk step if it is driven by non-sparse distributions like Gaussian.
The CIFAR10 examples, although less obvious than the MNIST examples, also show similar tendency---the $\alpha$-stable random walk with smaller $\alpha$
provides sparser adversarial patterns.

%
%
%
%
%
%

\section{Discussion}

Many defense strategies have been proposed to counter adversarial attacks \cite{madry2017towards, Pouya, SriArXiv18}. 
%
%
However, it happened many times that a new defense strategy is broken down by a newer attacking strategy only a few months after its proposal.
Thus, the adversarial defense problem has not been solved even on the toy MNIST data set,
although defense is considered much harder for larger data sets.

One recent finding in this ensuing arms race between new defense and attacking strategies
is the importance of the metric of the distortion, i.e., how to measure the distance from the original sample.
In whitebox attacks,
the $L_\infty$, $L_1$, and $L_2$ norms are often used to measure the distortion
\cite{Goodfellow, madry2017towards, Carlinib, chen2017ead}.
Interestingly, the state-of-the-art defense method proposed by Madry et al. \cite{madry2017towards} has shown to be robust against attacks with $L_\infty$ bounded perturbations,
while it has been found to be vulnerable against attacks with the elastic net ($L_1$ plus $L_2$) bounded perturbations \cite{chen2017ead, sharma2017ead}. 

Naturally, the choice of the perturbation metric impacts the human perception, e.g.,
limiting the $L_2$ norm makes the visual quality of the image better while limiting the $L_1$ norm assures sparsity of the distortion. 
However, the finding above implies that sparser regularization might help gradient-based whitebox optimization find stronger adversarial samples.
Our results in this paper might imply something similar or at least related---sparser random walk steps help exploration move along the decision boundaries,
and produce stronger adversarial samples under the blackbox scenario.
Further investigation is left as future work.

\section{Conclusion}


In this paper, we investigated how statistics of random variables affect random walk based blackbox attacking strategy.
Specifically, we proposed   \method{}, a generalization of the state-of-the-art {boundary attack}, 
where random walk is driven by symmetric $\alpha$-stable random variables.
Our experiments showed that
 the impulsive characteristics of the $\alpha$-stable distribution enables efficient exploration in the data space without crossing decision boundaries,
 producing stronger adversarial samples.
In our future work, we investigate the use of explanation methods \cite{bach2015pixel, MonDSP18, SamITU18b, LapNCOMM19} for adversarial attack detection and further study the relation between norm bounds, sparse exploration, and the quality of adversarial samples.


\bibliographystyle{unsrt}
\bibliography{eusipco}

\end{document}